\documentclass[journal]{IEEEtran}
\usepackage{algpseudocode}

\usepackage{color, soul}
\usepackage{multirow}
\usepackage{amsmath}
\usepackage{graphicx}
\usepackage{caption}
\usepackage{subfigure}
\usepackage{cite}
\usepackage[utf8]{inputenc}
\usepackage[table]{xcolor}
\usepackage{breqn}
\usepackage[ruled,vlined]{algorithm2e}
\usepackage{amsfonts}
\usepackage{amsthm}

\usepackage[export]{adjustbox}
\usepackage{hyperref}
\usepackage{cleveref}
\usepackage{xcolor}
\usepackage{soul}

\usepackage[disable]{todonotes}

\begin{document}
\title{Unbalanced Fingerprint Classification for Hybrid Fingerprint Orientation Maps}

\author{Ravi~Prakash, ~\IEEEmembership{Student Member,~IEEE} and~Sinnu~Susan~Thomas,~\IEEEmembership{Member,~IEEE}

\thanks{Ravi Prakash and Sinnu S. Thomas are with the School of Computer Science and Engineering, Digital University Kerala ( formerly IIITMK) Kerala 695317 India. e-mail: ravi.csres22, sinnu.thomas@duk.ac.in}
\thanks{Source code is made available on request.}
}


\maketitle
\begin{abstract}
This paper introduces a novel fingerprint classification technique based on a multi-layered fuzzy logic classifier. We target the cause of missed detection by identifying the fingerprints at an early stage among dry, standard, and wet. Scanned images are classified based on clarity correlated with the proposed feature points. We also propose a novel adaptive algorithm based on eigenvector space for generating new samples to overcome the multiclass imbalance. Proposed methods improve the performance of ensemble learners. It was also found that the new approach performs better than the neural-network based classification methods. Early-stage improvements give a suitable dataset for fingerprint detection models. Leveraging the novel classifier, the best set of `standard' labelled fingerprints is used to generate a unique hybrid fingerprint orientation map (HFOM). We introduce a novel min-rotate max-flow optimization method inspired by the min-cut max-flow algorithm. The unique properties of HFOM generation introduce a new use case for biometric data protection by using HFOM as a virtual proxy of fingerprints.
\end{abstract}


\IEEEpeerreviewmaketitle

\section{Introduction}

Biometric data is one of the best methods for uniquely identifying individuals \cite{9056799_biometric_app}. Among many biometric information, the three most common types include - facial recognition, iris scan, and fingerprints. The iris scan is mainly preferred for children over other options as, irrespective of much other biometric information, it remains unchanged throughout life. However, due to many other problems associated with iris scanning, it is not as widely used as fingerprints. On the other hand, recording fingerprints is quick, easy, and economical among these three approaches. Research has proven that fingerprint biometrics can be a suitable method for identifying any individual \cite{9364994}. Moreover, due to being one of the earlier methods of non-repudiation, the study of fingerprints holds a significant place in biometrics \cite{JAIN201680}.

\subsection{Motivation and Challenges}

Now, when fingerprint sensors are prevalent methods for authentication, the chances of stealing one's fingerprint seem much more manageable. The success of fingerprint-based identification lies in efficiently extracting the fingerprint and its features (like minutia points). The cause of incorrect minutiae point detection can be the noise, an inappropriate orientation of finger(s) while providing the scan, or dull fingerprint recording due to the noisy scanner. Several reasons behind these problems include a need for more knowledge to take a correct fingerprint scan at home using an impression pad, often done to self-attest the documents. Sometimes, previous fingerprint impressions and dust lead to unnecessary noise while scanning the fingerprints since the scanners often stay exposed to the environment. The biometrics are also used for data protection through encryption, similar to alphanumeric passwords. However, unlike passwords, biometric information can not be changed. Therefore, it has been found that fingerprints are better in terms of security over passwords \cite{zimmermann2020password}.

This paper primarily centers on detecting poorly scanned fingerprint images based on their visual clarity through dataset balancing, image comparison, and using machine learning methods for classification. We consider the threat of misusing fingerprints to address the problem associated with them. We address the creation and selection of the most optimal features for the efficient classification of images. As a result, this paper also proposes the development of a virtual proxy for fingerprints, HFOM, that can ensure the security of fingerprints, as the reconstruction of original fingerprints is not possible using the same. Hence, this resulting HFOM can also be referred to as an image hash of fingerprints.

\subsection{Paper Organization}
This paper explores fingerprint biometrics and leverages the quality of its uniqueness for dynamic applications through HFOM. 
The main contribution of this paper is as follows:
\begin{itemize}
    \item Proposed novel features for better fingerprint classification. 
    \item Proposed UC-FLEM, i.e., an efficient method for fingerprint classification using fuzzy logic with a multi-layered architecture.
    \item Proposed a method to overcome the dataset imbalance and improve the performance of ensemble learner classifiers.
    \item Built an approach to generate a one-way method HFOM using a set of multiple fingerprints. 
    \item Finally, we make our code available to the community.
\end{itemize}

The remainder of this article is as follows: Section \ref{sec:litreview} shows related work in the literature. Section \ref{sec:methodology} presents the proposed novel fingerprint detection and classification method. A step-wise approach for new feature creation and overcoming the dataset imbalance is provided in the same section. It also explains the HFOM generation process. Along with the results, a comparative analysis among proposed and several existing methods is reported in Section \ref{sec:results}, followed by conclusions in Section \ref{sec:conclusion}.

\section{Literature Review}
\label{sec:litreview}

In the last few years, there has been ample research {\cite{manickam2023retraction, YUN2006101, manickam2019score}} in the domain of fingerprint biometrics \cite{tu2020fingerprint, JAIN201680, 10.1109/ICASSP.2016.7472049, 7471822, 7952518}. Many researchers have developed algorithms for detecting, clustering, and enhancing fingerprints. Latent-fingerprint analysis \cite{shreya2023gan, 5456043, 9053801} was a tedious process used for over a hundred years to detect criminal fingerprints better. These scans often have localization errors \cite{5456043} for one Region of Confidence \cite{1687-5281}. It is challenging to get the variance in pixels of small block size while analyzing the image in its locality \cite{1687-5281}. 

A labelled dataset is required in the process of classification for training. A smaller and unbalanced dataset can not develop any reliable classification model in such a situation. While seeking the labelled datasets, the issue is their size and balance among the data points. Methods such as SMOTE \cite{10.1613}, KMeans SMOTE \cite{2018.06.056}, ADASYN \cite{4633969}, and AdaSS \cite{10.1111/j.1468-0394.2010.00513.x} are used for balancing dataset by sample generation and eigenvalue thresholding \cite{80959}. SMOTE is non-adaptive, leading the samples to be dominant towards one (or more) specific feature and often generates overlapping samples. Although the AdaSS method is used in \cite{8489572} to tackle the multiclass imbalance, it still removes some points from the most dominant class. The proposed method must generate samples to define separation among the classes. 

Fuzzy logic is used for fingerprint description, retrieval, and identification \cite{momaninovel}. The descriptors are developed based on the texture of the fingerprint images by studying it partially \cite{10.1007/978-3-642-23713-3_29, 8461885}. The final texture is defined based on the overall image analysis, where the most dominant texture is considered. Arjona \textit{et al.} \cite{10.1007/978-3-319-19324-3_14} proposed a fuzzy-logic based system for fingerprint retrieval. This inspired us to work in the same direction to classify fingerprint images. A fuzzy fingerprint identification method \cite{10.15837/ijccc.2010.4.2510} works by matching two images comparatively.  

Fingerprint biometrics are used in several application areas, such as education \cite{okereafor2020fingerprint} and security \cite{yang2021biometrics}. The recognition of fingerprints \cite{chowdhury2022contactless} for authentication \cite{6044711_fp_App}, \cite{castro2023medical}, and data encryption are the major applications behind the same. While reviewing the literature, we also came across the relevant work on synthetic fingerprint generation \cite{9506386_fpGen}, inspiring our investigation into novel methods for generating hybrid fingerprint derivatives, suitable for security applications. 

Hybrid image generation aims to target a high similarity from the original ones to make the generated images more realistic. Structure Similarity Index Method (SSIM) \cite{nilsson2020understanding, wang2004image_SSIM_main, bakurov2022structural_SSIM_ups} is one of the frequently used methods that aims to quantify the similarity between any two images. Several other scoring methods like PSNR \cite{5596999_SSIM_FSIM}, FSIM \cite{5596999_SSIM_FSIM, ding2020image_NEW_meth}, and Complex Wavelet Structural Similarity (CWSS) \cite{sampat2009complex_SSIM_etc} are also found to be useful as per the literature on the image similarity. 

Existing literature must concentrate on fingerprint classification based on dry/wet/standard visual clarity. While working on the imbalanced dataset, we propose UC-FLEM as a fuzzy-logic-based classifier to address this issue. Irrespective of many state-of-the-art models that divide and analyze the images into parts, we collectively split the process into two phases and four layers. The UC-FLEM approach is followed after a more accurate feature extraction process, further boosting the classification model. Later, we utilized this model to get the best set of fingerprints and generate a hybrid fingerprint orientation, HFOM. HFOM is an image generated using a one-way function and is further used to propose an asymmetric data encryption/decryption algorithm. Once HFOM is generated, the original fingerprints can not be retrieved back, making HFOM a good substitute for fingerprint biometrics.

\section{Methodology}
\label{sec:methodology}

\subsection{Fingerprint Classification}
\label{subsec:uc-flem}
\begin{figure}[ht]
    \includegraphics[width=\linewidth]{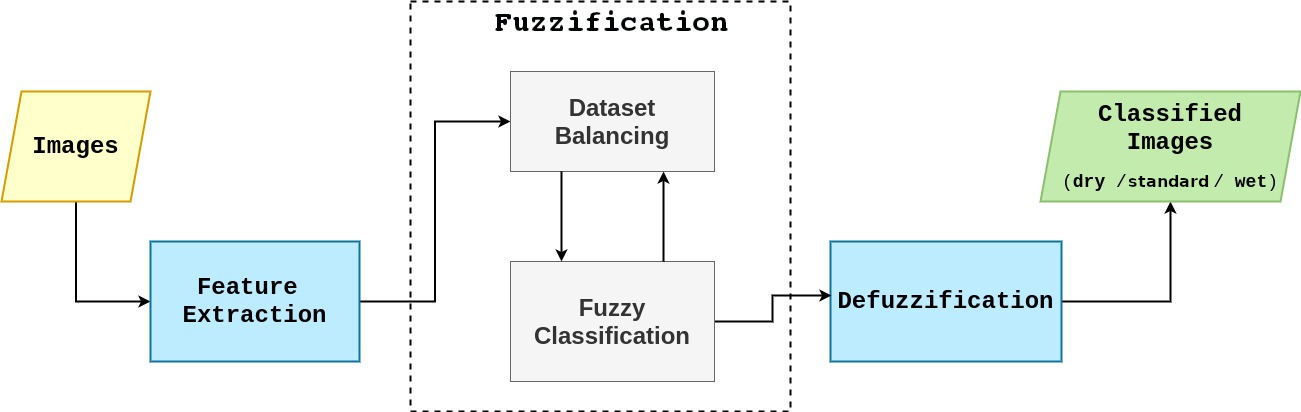}
    \centering                                     
    \captionsetup{justification=centering}
    \caption{Proposed Approach.}
    \label{fig:basicFlow}
\end{figure}

We propose the fingerprint classification as shown in Fig. \ref{fig:basicFlow}. 

We consider the FVC2000 fingerprint database \cite{fvc2000} to prepare the dataset $D$. The system takes grayscale fingerprints and prepares the dataset by extracting the features. The feature dataset is passed for fuzzification in two phases. After that, the defuzzification rules are applied to the outcomes for the final classification results.

\subsubsection{Feature Extraction}
We calculate mean $\mu$\cite{YUN2006101}, variance $\sigma ^ {2}$ \cite{YUN2006101} and ridge to valley thickness ratio $RVR$ \cite{YUN2006101} using the covariance matrix from the images. We calculate the block directional difference ($BDD$) while dividing the input image into multiple blocks $B$ of size $3\times3$ along with global padding of $1$ pixel, keeping uniformity in block splitting. We calculate a block $B$ (Eq. \ref{eq:blockFPfeature}) having pixels $P_{ij}$, where $i$ and $j$ represent the rows and columns of the image, respectively. A $9\times9$ mask is used to generate the slit sum $S_{i}$ (Eq. \ref{eq:slitsum}) for the center pixel of each block $B_{j}$ taking one block at a time. The $BDD$ for the current block's center pixel is the difference between the maximum and minimum $S_i$. We then compute the orientation change while convolving the positive Laplacian Filter $L$ with each $B$ to find the gradient $G_{ij}$ for the blocks $B_{ij}$ at $i^{th}$ row and $j^{th}$ column. The orientation change is the arctan of $G_{ij}$.

\begin{equation}
   B = ( \,\: \sum_{i=1}^{3}\sum_{j=1}^{3} P_{ij} \: ) \, - P_{22}.
    \label{eq:blockFPfeature}
\end{equation}

\begin{equation}
    S_{i} = \sum_{j=1}^{8} B_{j}
    \label{eq:slitsum}
\end{equation}

As many of the above features are multi-valued, we compute a single-valued variant by taking the average for $\theta$ = $\theta_{avg}$, for $BDD$ = $BDD_{avg}$ and for $RVR$ = $RVR_{avg}$. Since the $RVR_{avg}$ is found to be very low for the images, we compute the squared sum of $RVR$ as $SSRVR$ (Eq. \ref{eq:SSRVR}) that gives the collective measure of the ridge to valley thickness. The computational error $\epsilon$ is considered as $10^{16}$.
\begin{equation}
    SSRVR(i,j) = \sum_{i=1}^{n}\sum_{j=1}^{n} (RVR_{ij} \times \epsilon)^{2}
    \label{eq:SSRVR}
\end{equation}

We extract these features for the accurate classification of the fingerprints. 

\subsubsection{Fuzzy Classifier}
\begin{figure}[ht]
    \includegraphics[width=\linewidth]{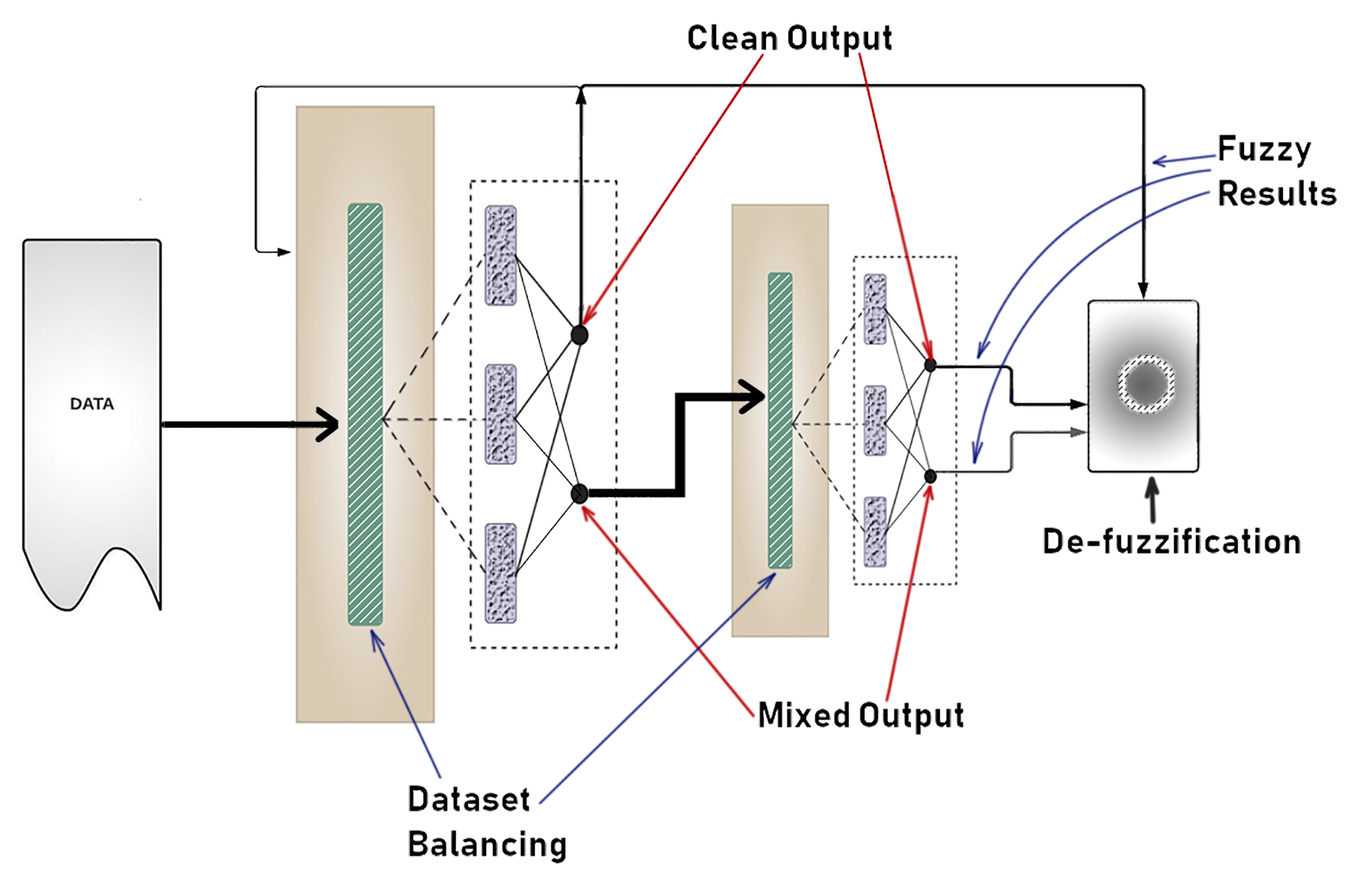}
    \centering
    \captionsetup{justification=centering}
    \caption{Dual-phase Dual-layered Architecture of Fuzzy Classifier.}
    \label{fig:flowChart}
\end{figure}

We classify the images using the proposed features. In order to further enhance the classification performance, we propose a fuzzy logic-based classifier. It is a dual-phase process working on two different types of classifiers. We consider Random Forest Classifier \cite{10.1023} and Gradient Boosting \cite{10.1214/aos/1013203451} for fuzzification.
The detailed working of the fuzzy classifier begins with balancing feature data that is later classified in a fuzzy manner over the different layers of two phases. The entire internal process is shown in Fig. \ref{fig:flowChart}.

\subsubsection{Phase}
Each phase is further broken down into two layers, having separate classifiers, i.e., $\Psi_1$ and $\Psi_2$. It is based on the predictions of every data point $x \in D$ over both of the layers, i.e., $L_{1}$, $L_{2}$ (Eq. \ref{eq:phaseEval}). The data is backed up once the data processing and partial classification are over at a phase. Among the phase-1 results, only a subset of image data points is transferred to the next phase $\tau$, and the remaining output is considered the final fuzzy results $\rho$. It is decided based on internal processing done at every layer. All of the $\rho$ are collectively taken for defuzzification.

\begin{equation}
    \left.
    	\begin{array}{ll}
    		\rho = \{L_{1}\} \cap \{L_{2}\}  & \mbox{if } L_{1} \cap L_{2} \ne \phi \\
    		\tau = \{D\} - \{\rho\} & \mbox{if }  \text{otherwise}
    	\end{array}
     \right.
    \label{eq:phaseEval}
\end{equation}

\subsubsection{Layers}
\label{subsec:layers}
Each of the two layers is dedicated to specific roles triggered sequentially. At every layer, a separate classifier is used for splitting the predictions. 

Layer-1 focuses on balancing $D$ by generating samples using an adaptive algorithm, by creating a $n \times n$ matrix $\mu$ for each class $\overline{C}$ with labels $c$. We consider $M$ is the medoid and $C_1$, .. $C_{n-1}$ as the $n-1$ closest points to the $M$. Here, $n$ is the number of features we extracted earlier as in Eq. \ref{eq:featuresMat}. The eigenvector space is computed for $\mu$, and the sample points are generated at the largest eigenvalue. We use Eq. \ref{eq:verifyK} and \ref{eq:verifyK0} to maintain the previous ($\kappa$) and latest ($\kappa_o$) KL-Divergence of the class distribution and avoid any replication in a balanced dataset $D_o$. The best models are selected for Layer-1 and Layer-2 using  Algorithm \ref{alg:phase}, and initial predictions are made by $\Psi_{1}$ as $L_{1}$. Based on $L_{1}$, $D$ is divided into three subsets and supplied to the next layer for fuzzification. Layer 2 takes data fragments and a classification model $\Psi_{2}$ from Layer 1. Classification is done at this layer as $L_{2}$, and overall results are grouped for the phase as given in Eq. \ref{eq:phaseEval} as $\rho$ and $\tau$. The value $\tau$ is supplied to phase 2 as a testing dataset. 
\begin{equation}
    \mu = 
    \begin{bmatrix}
            C_{1} & C_{2} & C_{3} & .... & C_{n-1} & M
    \end{bmatrix}^T
    \label{eq:featuresMat}
\end{equation}

\begin{equation}
    \kappa = \{max(\mathbb{KL}(\overline{C} \| D))\} \cap \{c\}
    \label{eq:verifyK}
\end{equation}

\begin{equation}
    \kappa_{o} = \{max(\mathbb{KL}(\overline{C} \| D_o))\} \cap \{c\}
    \label{eq:verifyK0}
\end{equation}

In the end, $\tau$ from phase-1 and $\rho$ from both phases are grouped as the fuzzified results. Final results are computed after defuzzification. A standard method used in both phases is shown in Algorithm: \ref{alg:phase}.

\begin{algorithm}[ht]
\KwIn{Training dataset $T$, test dataset $D$}
\KwOut{Accuracy of model used on $i^{th}$ layer $A_{i}$ and classified output $O$}
$A_{i} \gets \text{accuracy(train}(\Psi_{i},T)) ~~ \forall~i \in \{1,2\}$\;

\begin{equation}
    \Lambda_{1} = 
    \left\{
    	\begin{array}{ll}
    		\Psi_{1}  & \mbox{if } A_{1} > A_{2} \\
    		\Psi_{2} & \mbox{if } A_{1} \leq A_{2}
    	\end{array}
    \right.
    \label{eq:layer1Model}
\end{equation}

\begin{equation}
    \Lambda_{2} = 
    \left\{
    	\begin{array}{ll}
    		\Psi_{1}  & \mbox{if } A_{1} < A_{2} \\
    		\Psi_{2} & \mbox{if } A_{1} \geq A_{2}
    	\end{array}
    \right.
    \label{eq:layer2Model}
\end{equation}

$R_{i} \gets \text{classify}(\Lambda_{i}, D) ~~ \forall~i \in \{1,2\}$\;

$O \gets \bigcup\limits_{i = 1}^{2} R_{i}$\;
$A_{i} \gets \text{accuracy}(\Lambda_{i}) ~~ \forall~i \in \{1,2\}$\;

\Return $A_{i}, O$.
\caption{Basic Fuzzification on each Phase.}
\label{alg:phase}
\end{algorithm}

\subsubsection{Defuzzification}
Fuzzified results from phase-1 ($\rho_{1}$) represent fingerprints with maximum chances to be dry/standard/wet, and those from phase-2 ($\rho_{2}$) represent a highly probable fingerprint as dry/standard/wet. Hence, a basic voting rule is considered for defuzzification of $\rho_{1}$ and $\rho_{2}$. In the case of $\tau$, the voting rule fails due to the ties among the classification done at different layers. Therefore, based on the accuracy, the best of the two models is selected as $\Psi_{o}$ as in Eq. \ref{eq:defuzzRules}.
\begin{equation}
    \left.
    	\begin{array}{ll}
        	\Lambda_{\alpha{i}} = \frac{\sum_{i=1}^{2} A_{il_j}}{2} ~~ \forall j \in \{1,2\} \\
            \Psi_o = max(\Lambda_{\alpha{i}}) ~~ \forall i \in \{1,2\}.
    	\end{array}
     \right.
    \label{eq:defuzzRules}
\end{equation}

\subsubsection{Dataset Balancing}
After preparing the feature dataset, it was found that there was an imbalance among the three classes. Due to the smaller dataset, the application of AdaSS \cite{8489572} and undersampling are not feasible. However, it inspired us to tackle this problem by exploiting boundary region competencies of the data points falling in any specific class. For every class, we took the medoids, and based on the Euclidean distance, the medoid and its five closest points were chosen to form a $6\times6$ matrix using Eq. \ref{eq:featuresMat}. The direction is computed for new sample points using the eigenvector space and the most dominant eigenvalue. The introduced mechanism generates points in that direction from where more information could not be learned. 

The new sample is only added to the dataset if it maintains the class distribution (Eq. \ref{eq:verifyK}, \ref{eq:verifyK0}). Due to being adaptive, every sample also generates a newer one, so the whole process is repeated.

\subsection{Fingerprint Hybridization and Hybrid Fingerprint Orientation Map (HFOM) Generation}  
\label{subsec:hfom}
\begin{figure}
    \includegraphics[width=\linewidth]{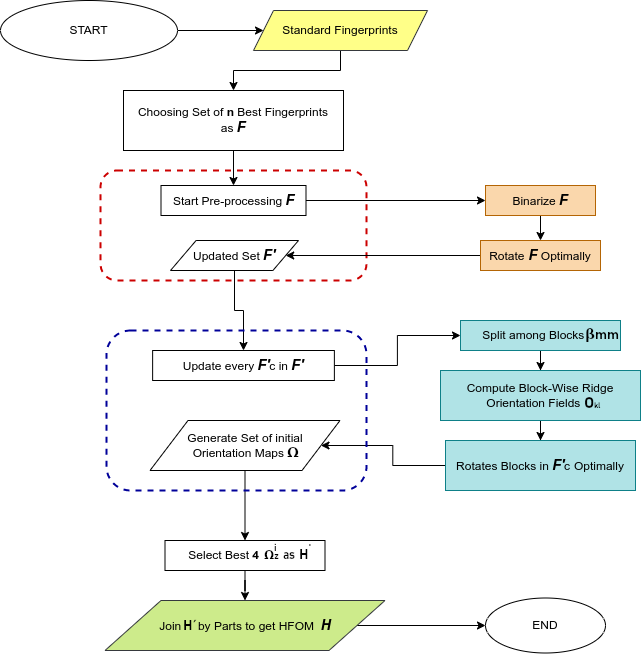}
    \centering
    \captionsetup{justification=centering}
    \caption{Creating HFOM from Pool of Classified Fingerprints.}
    \label{fig:hybridFlowChart}
\end{figure}

The proposed hybrid fingerprint orientation map is a unique derivative of input fingerprints. The overview of HFOM generation is given in Fig. \ref{fig:hybridFlowChart}.

\subsubsection{Image Pre-processing}  
The fingerprints are prepossessed to make them compatible with the HFOM generation. Initial transformation is done in terms of dimensions and color intensity. The model crops and resizes every fingerprint image to a size of resolution $s \times s$ pixels. 
The given method is suitable for grayscale images only. 

After the initial prepossessing, each fingerprint image $F$ is classified among dry/standard/wet using the proposed UC-FLEM model. All the $standard$ fingerprints are extracted to select a set of best $n$ fingerprints (Eq. \ref{eq:fpSet}). This selection of $F$ is made after sorting all fingerprints in increasing order of features $\mu$, $\sigma ^ {2}$ and $RVR$ respectively. The final set of $F_c \in F$ contributes to the generation of HFOM. 

\begin{equation}
    F = \{ F_{c} \} ~~ \forall ~ c \in \{0,1,...,n-1\}
    \label{eq:fpSet}
\end{equation}

\subsubsection{Optimal Fingerprint Orientation}
The process of generating HFOM using $F$ is divided into three steps. The first part involves the generation of optimal fingerprint orientation fields for each of the $n$ fingerprints. The initial step is binarization, where for every fingerprint $F_c \in F$, the ${P_{ij}}^c$ is changed. Here ${P_{ij}}^c$ represents the value of the pixel at the row $i$ and column $j$ in image $c$ of the $F$. 
The final set of binarized $F$ consists of $F'_c \in F'$ as in Eq. \ref{eq:fpBinSet}. Due to binarization, $F'_c$ has a lower color variance than the respective $F_c$. 

\begin{equation}
    F' = \{ F'_{c} \} ~~ \forall ~ c \in \{0,1,...,n-1\}
    \label{eq:fpBinSet}
\end{equation}

Now the total count of common pixel values ${P'_{ij}}^c$ for each of the images $F'_{c}$ in the stack of the image $F'$ is computed using Algorithm \ref{alg:partialFOM}. It is followed by maximizing the common pixel count throughout the $F'$. Hence, every $F'_c \in F'$ is rotated by an angle $\theta_{c} \in \theta$ (as in Eq. \ref{eq:rotAngles}) using a new method of optimization named min-rotate max-flow. It is inspired by the normalized cut method min-cut max-flow \cite{868688_normCut}. The aim is to achieve a uniform value of ${P'_{ij}}^c$ at position $(i,j)$ for every $F'_{c}$ by keeping the $\theta_{c}$ minimum. All of the intermediate rotations are only momentary, and those are undone if the rotation does not increase the number of ${P'_{ij}}^c$, which is computed as per the Algorithm \ref{alg:partialFOM}.

\SetKwComment{Comment}{/* }{ */}
\begin{algorithm}[ht]
\caption{Count Total Number of Common ${P'_{ij}}^c$ in $F'$}\label{alg:partialFOM}
\KwIn{A stack of $n$ fingerprints as $F'$}
\KwOut{Number of pixels with common values as $P_{count}$}

$c \gets \{1,2, ..., n-1 \}$ \; 
$c' \gets 0$ \; 
$\xi \gets \emptyset$ \Comment*[r]{\text{initializing to count} ${P'_{ij}}^c$}

\While{$c' \neq n$}{
    $i,j \in  \{ 0, 1, 2, ...,s'-1 \}$ \; 
    $P_{ij}^{c'} \in  F'_{c'}$ \;
    \begin{equation}
        \xi_{ij} =
        \left\{
        	\begin{array}{ll}
        		1  & \mbox{if } P_{ij}^0 = P_{ij}^{c_\circ} \\
        		0  & \mbox{if } P_{ij}^0 \neq P_{ij}^{c_\circ}
        	\end{array}
        \right.
        \forall ~ {c_\circ} \in c-\{0\}
        \label{eq:pxlValCheck}
    \end{equation}
    $c' \gets c' + 1$ \; 
}

\begin{equation}
   P_{count} = ( \,\: \sum_{i=0}^{s}\sum_{j=0}^{s} \xi_{ij} \: )
    \label{eq:pxlCounter}
\end{equation}

\Return $P_{count}$
\end{algorithm}

\begin{equation}
    \theta = \{0^\circ, 90^\circ, 180^\circ, 270^\circ\}
    \label{eq:rotAngles}
\end{equation}

It is an iterative process where the image orientation is updated for the whole $F'$ repeatedly as long as the increment in the number of common ${P'_{ij}}^c$ is noticed. In the optimal condition, no increment is noticed in the joint pixel counts with or without any orientation change for $F'_c$.

\subsubsection{Ridge Orientation Fields Generation}
The stack of binarized images oriented at the optimal angle(s) is considered for the initial orientation map generation for each $F'_{c}$. At this stage we consider splitting the complete image in a matrix of $m \times m$ blocks $\beta_{mm}$ (Eq. \ref{eq:blockNum}) of size $b_{s} \times b_{s}$ pixels, across horizontal and vertical sides such that $b_{s} / 2 \in \mathbb{W}$ i.e. $b_{s}$ is an odd number. In order to club every pixel in some block, all of the $\beta_{mm}$ must be uniformly distributed throughout the $F'_{c}$. Hence, a padding $p_{\circ}$ is added at the edges of every $F'_{c} \in F'$. Here, the padding is a sequence of white pixels having an overall thickness of $p_{\circ}$ pixel throughout $F'_c$. The condition in Eq. \ref{eq:paddingCount} must be followed while choosing the value $b_{s}$ in order to minimize $p_{\circ}$. Therefore, it increases the size, i.e., pixel resolution of $F'_{c}$ from $s \times s$ to $s' \times s'$ (Eq. \ref{eq:paddedSize}). $p_{\circ}$ is equally distributed across all of the four sides of $F'_c$. However if $p_{\circ} / 4 < 1$, then $right$ and $bottom$ sides are left unpadded. For every block $\beta_{kl} \in F'_c$ selected from $F'$, an orientation field is generated that represents the overall direction for the flow of ridges in it. The orientation field is generated for a given block $\beta_{kl}$ as per Eq. \ref{eq:getOrtFldAngle} in Algorithm \ref{alg:ortMapCalc}.

\begin{equation}
   \beta_{mm} = \{ \beta_{kl} \} ~~
   \forall ~ k,l \in \{0, 1, 2, ..., m-1\}
    \label{eq:blockNum}
\end{equation}

\begin{equation}
   p_{\circ} = (\lceil \frac{s}{b_{s}} \rceil \times b_{s}) - s ~~
   \forall ~ p_{\circ} < \frac{b_{s}}{2}
    \label{eq:paddingCount}
\end{equation}

\begin{equation}
    s' = s + \frac{p_{\circ}}{2}
    \label{eq:paddedSize}
\end{equation}

\begin{figure}
    \includegraphics[width=\linewidth]{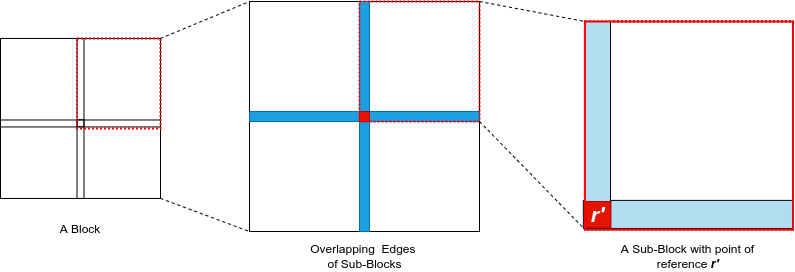}
    \centering
    \captionsetup{justification=centering}
    \caption{Block $\beta_{kl}$ as a set of overlapping sub-blocks $q_{\eta}$.}
    \label{fig:blockStruct}
\end{figure}

To get the ridge orientation field for any block present on row $k$ and column $l$, Eq. \ref{eq:subBlockOrt} is used to splitting $\beta_{kl}$ further into four identical sub-blocks $q_\eta \in Q_{kl}$. All of the $q_\eta$ have an overlapping of width 1-pixel on both horizontal and vertical inner edges as shown in Fig. \ref{fig:blockStruct}.

\begin{equation}
   Q_{kl} = \,\: \{ q_\eta: {\beta_{kl}}^\eta \}\: ~~
   \forall ~ \eta \in \{ 1,2,3,4 \}
    \label{eq:subBlockOrt}
\end{equation}

As there is an overlapping among all $q_\eta$, the center pixel of the block $\beta_{kl}$ is treated as the point of reference $r'$. This process is followed by a necessary rotation of pixels $P_{xy}$ located at the position $(x,y)$ to location $(x', y')$ by the factor $\alpha \in \theta$ (Eq. \ref{eq:subBlockRot}) on each $q_\eta$. Once rotated, the partial ridge orientation fields $O'_{kl}$ are computed for each $q_\eta$ and rotated back to the original position by the angle of $-\alpha$. 

\begin{equation}
\left.
   \begin{array}{ll}
     \begin{bmatrix}
      x'\\ 
      y'
    \end{bmatrix} =  \begin{bmatrix}
      cos\gamma & -sin\gamma\\ 
      sin\gamma	& cos\gamma
    \end{bmatrix}  \begin{bmatrix}
      x\\ 
      y
    \end{bmatrix}
    ~~ \forall ~ \gamma ~ \exists ~ \alpha \in \theta \\
    ~~~~ \text{such ~ that} ~ \gamma \in \{ \alpha, -\alpha\}
    \end{array}
    \right.
    \label{eq:subBlockRot}
\end{equation}

\begin{figure}
    \includegraphics[width=\linewidth]{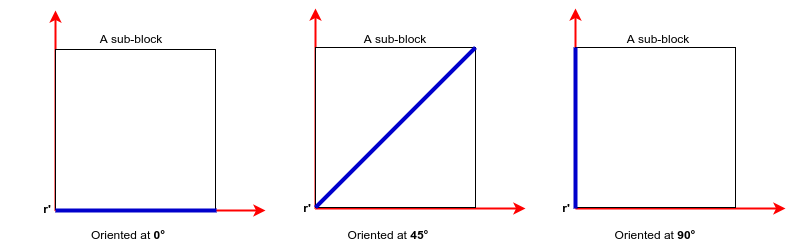}
    \centering
    \captionsetup{justification=centering}
    \caption{Three possible orientation fields in $q_{\eta}$ having origin at $r'$.}
    \label{fig:3subBlockOrt}
\end{figure}

In $q_\eta$, there can only be one possible orientation line out of the three, i.e., oriented at angle $\theta' = \{0^\circ, 45^\circ, 90^\circ\}$ as shown in Fig. \ref{fig:3subBlockOrt}. Finally, $O'_{kl}$ from either $2$ or $3$ sub-blocks are combined altogether to generate the final orientation field $O_{kl}$ for current $\beta_{kl}$ as per Algorithm \ref{alg:ortMapCalc}.

\begin{algorithm}[ht]
\KwIn{Set of sub-blocks $q_\eta$}
\KwOut{Ridge Orientation Map for given block $O_{kl}$}


\For{\text{iterate} $\eta \in \{1,2,3,4\}$}{
    \begin{equation}
       \left.
       \begin{array}{ll}
            \alpha' = \text{orientation ~ flow ~ of ~ black ~ pixels} ~ \in q_\eta \\
            ~~~~~~~ \forall ~ \alpha' \in \theta' \\
            {O_{kl}}^\eta = \text{straight ~ line} ~ at ~ \alpha'
        \end{array}
        \right.
        \label{eq:getOrtFldAngle}
    \end{equation}
}

\begin{equation}
   O_{kl} = \,\: \emptyset \cup \{{O_{kl}}^\eta\}
    \label{eq:ridgeOrtField}
\end{equation}

\Return $O_{kl}$.
\caption{Ridge Orientation Map for $\beta_{kl}$.}
\label{alg:ortMapCalc}
\end{algorithm}

\subsubsection{HFOM Generation}
The collective representation of ridge orientation fields for all of the blocks $\beta_{mm} \in F'_{c}$ as an image is the desired fingerprint orientation map $\Omega_{c} \in \Omega$ of $F'_{c}$. Once we have $\Omega$, each $\Omega_{c}$ is divided among four equal and non-overlapping blocks ${\Omega_{c}}^\eta$. 

\begin{equation}
   \Omega = \,\: \{ \Omega_{c} \} \: ~~
   \forall ~ c \in \{ 0,1,...,n-1 \}
    \label{eq:ortMapSet}
\end{equation}

Finally a set of the four ${\Omega_{c}}^\iota$ is selected i.e. ${H_{z}}$. ${\Omega_{c}}^\iota$  are selected such that no two $\iota$ and $z$ have common values. Finally, the desired HFOM $H$ is generated by joining all four ${H_z}$.

\begin{equation}
   \left.
   \begin{array}{ll}
        H_{z} = \{ {\Omega_{z}}^\iota \} ~~ \forall ~ z \in c, ~~ \iota \in \eta \\
        \exists ~ \iota_\circ \neq \iota', ~ z_\circ \neq z' ~~ \forall ~ \iota_\circ, \iota' \in \iota, ~~ z_\circ, z' \in z.
    \end{array}
    \right.
    \label{eq:HFOMbyParts}
\end{equation}

\begin{equation}
    H = \emptyset \cup H' ~~ \forall ~~ H' \subset H_{z}
    \label{eq:finalHFOM}
\end{equation}

\section{Experimental Results}
\label{sec:results}
\subsection{Fingerprint Classification}
\subsubsection{Dataset} 
FVC2000 is a widely used database in fingerprint enhancement \cite{9206836, 6912970_tifs}. We prepared the dataset of $800$ fingerprints by taking images from FVC2000\_DB4\_B \cite{fvc2000}. The image resolution is $160\times160$ pixels, and the dataset is manually tagged for classification. The dataset used for training and testing images is in the $70:30$ ratio. 

\subsubsection{New Features}  
Table \ref{tab:nonSmote} gives the model's performance using new features. Here $\mu$, $\sigma ^ {2}$, $SSRVR$, $BDD_{avg}$, $RVR_{avg}$, and $\theta_{avg}$ fit best for the desired classification.

\begin{table}[h]
\centering
    \begin{tabular}{|l|l|l|l|l|}
    \hline
    \textbf{Model}  &  {Accuracy}  &  {Recall}  &  {Precision}  &  {F1}   \\
    \hline \hline
    \multicolumn{1}{|l|}{Random Forest} &  79.2\%  &  78.3\%  &  78.3\%  &  78.2\%    \\
    \multicolumn{1}{|l|}{Gradient Boost}  &  77.5\% &  76.5\%  &  76.5\%  &  76.4\%  \\
    \multicolumn{1}{|l|}{Naive Bayes}  &  76.3\% &  73.5\%  &  76.6\%  &  76.0\%   \\
    \multicolumn{1}{|l|}{K Neighbors}  &  76.2\%  &  73.3\%  &  76.6\%  &  76.7\%   \\
    \multicolumn{1}{|l|}{Decision Tree}  &  73.2\%  &  71.3\%  &  73.8\%  &  73.1\%   \\
    \multicolumn{1}{|l|}{Linear SVM}  &  70.5\%  &  67.8\%  &  67.2\%  &  63.6\%   \\
    \hline
    \end{tabular}
    \caption{Comparing Performance of Different Classifiers for the Proposed Features with Unbalanced Dataset.}
    \label{tab:nonSmote}
\end{table}

The performance of Random Forest classifiers was the best, with a maximum accuracy of $79.2\%$, roughly $2\%$ more than Gradient Boost, as shown in Table \ref{tab:nonSmote}. Hence, we join these two classifiers for the fuzzification process, which performs better together.
      
\begin{figure}
    \includegraphics[width=\linewidth]{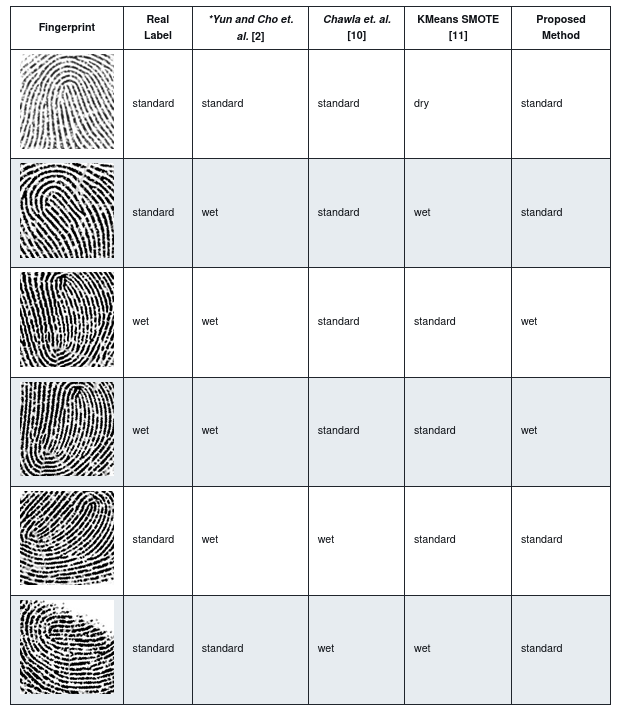}
    \centering
    \captionsetup{justification=centering}
    \caption{Classification Results (* uses the unbalanced dataset).}
    \label{fig:improvedNewBalancing}
\end{figure}

Table \ref{tab:threePerformance} shows the impact of using old features \cite{YUN2006101} with the proposed model. Images are classified using different algorithms \cite{10.1613,2018.06.056} and without-oversampling \cite{YUN2006101} as shown in Fig. \ref{fig:improvedNewBalancing}. The original labels are given in the second column. It consists of the images from the actual test set used during the model validation. The dry images consist of the vanishing ridge, while wet images have the vanishing valley regions. Images are only perfect for further usage with proper processing, and the proposed algorithm quickly sorts them out.

\subsubsection{Dataset Balancing}
We applied different oversampling methods along with the fuzzy logic-based classifier for fingerprint classification. It is seen in Table \ref{tab:threePerformance} that the proposed classification and data-balancing methods give the highest accuracy. The model can detect standard images out of dry and wet pools. The proposed method can detect fine details for any light or dark image. The proposed feature $SSRVR$ preserves the thin ridge and valley details. The overall performance of the proposed UC-FLEM approach is shown in Table \ref{tab:all_classifiers_Balanced}, where the proposed method is compared with different oversampling techniques.

\begin{table}[h]
\centering
     \begin{tabular}{ | m{1.8cm} | m{1.6cm} | m{1.6cm} | m{1.6cm} |} 
     \hline
     \textbf{Performance}  &  Yun and &  Chawla \textit{et} & Proposed \\
     \textbf{Measure}  & Cho \cite{YUN2006101} &  \textit{al.} \cite{10.1613} &  Method\\
     \hline \hline
         F1-Score  &  79.89\%&  82.45\%& 82.89\%\\
         Accuracy  &  80.42\%&  82.78\%& 83.15\% \\
         Precision  &  80.19\%&  82.70\%& 83.14\% \\
         Recall  &  79.96\%&  82.33\%&82.77\% \\
     \hline
 \end{tabular}
     \caption{Performance evaluation for resulting predictions.}
 \label{tab:threePerformance}
\end{table}
\begin{table}[h]
\centering
    \begin{tabular}{|l|l|l|l|l|}
    \hline
    \textbf{Proposed Fuzzy}  &  Accu-  &  Recall  &  Preci-  &  F1   \\
    \textbf{Classifier with}  &  racy\%  &  \%  &  sion\%  &  \%   \\
    \hline \hline
    \multicolumn{1}{|l|}{KMeans SMOTE \cite{2018.06.056}}  &  77.08 &  75.88  &  75.87  &  75.87  \\
    \multicolumn{1}{|l|}{SMOTE N}  &  77.92 &  76.94  &  76.95  &  76.84   \\
    \multicolumn{1}{|l|}{SVM SMOTE}  &  78.33  &  77.36  &  77.41  &  77.29   \\
    \multicolumn{1}{|l|}{SMOTE \cite{10.1613}}  &  82.78  &  82.33  &  82.70  &  82.45   \\
    \multicolumn{1}{|l|}{Proposed Method}  &  83.15  &  82.77  &  83.14  &  82.89    \\
    \hline
    \end{tabular}
    \caption{Performance with different oversampling methods.}
    \label{tab:all_classifiers_Balanced}
\end{table}

Comparing the performances across the phases indicates that KMeans SMOTE \cite{2018.06.056} and SMOTE are used in sampling for a small set of data points, whereas the proposed method gives a rich and balanced dataset in case of a large unbalance. A substantial improvement in the evaluation matrix (Table \ref{tab:threePerformance}) indicates that the proposed classifier and the proposed oversampling method exhibit outstanding performance.  

\vspace{1cm}

\subsubsection{Comparison with ANN Models} 
\label{subsub:CNN}

\begin{figure}
    \includegraphics[width=\linewidth]{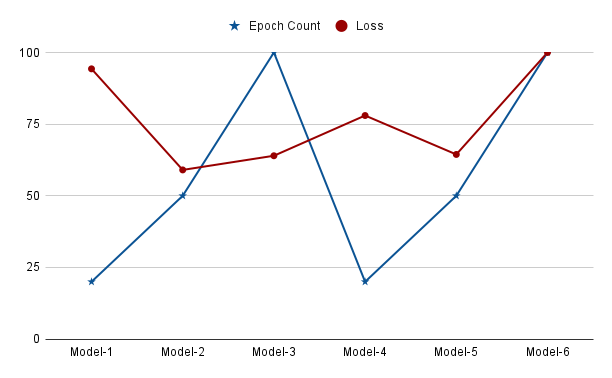}
    \centering
    \captionsetup{justification=centering}
    \caption{Information loss (in \%) of CNN models trained on raw (before binarization) dataset.}
    \label{fig:rawCNN_loss}
\end{figure}

\begin{figure}
    \includegraphics[width=\linewidth]{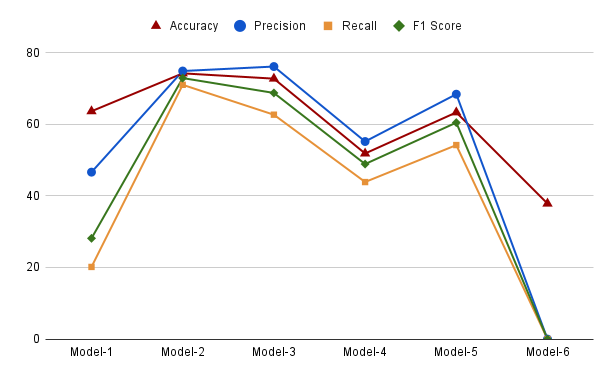}
    \centering
    \captionsetup{justification=centering}
    \caption{Performance (in \%) of CNN models trained on raw (before binarization) dataset.}
    \label{fig:rawCNN_prof}
\end{figure}

We compared the performance of the proposed classification system with that of the artificial neural network (ANN) based approach. We created a sequential CNN model with three convolution layers, each followed by a max-pooling layer. However, the three convolution layers varied with each other in terms of the features learned, i.e., (16, 32, 64), respectively. We tested another model variant by adding a fourth set of convolution and max-pooling layers, where the filter size was kept at $128$ during the convolution operation. These variants were later trained with three sets of epochs, i.e., 20, 50, 100. Fig. \ref{fig:rawCNN_loss} and \ref{fig:rawCNN_prof} show the information loss and performance measures for six different models trained on the raw \cite{fvc2000} fingerprint dataset, respectively.

\begin{figure}
    \includegraphics[width=\linewidth]{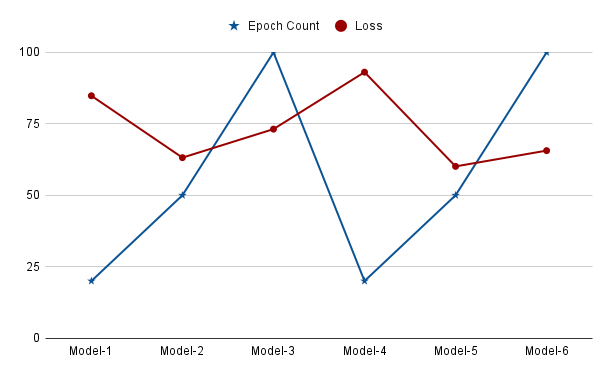}
    \centering
    \captionsetup{justification=centering}
    \caption{Information loss (in \%) of CNN models trained on the binarized fingerprint dataset.}
    \label{fig:binCNN_loss}
\end{figure}

\begin{figure}
    \includegraphics[width=\linewidth]{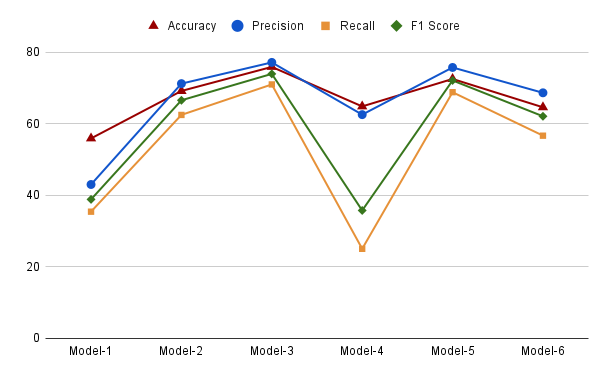}
    \centering
    \captionsetup{justification=centering}
    \caption{Performance (in \%) of CNN models trained on the binarized fingerprint dataset.}
    \label{fig:binCNN_prof}
\end{figure}

The model-2, with three convolution layers, performed best among all six and reflected an accuracy of $74.18\%$ and F1-score of $72.86\%$. Moving ahead, we trained these six CNN models (with the same set of configurations and architecture on the images processed based on the proposed binarization technique. After testing these models, the performance evaluations are visualized in Fig. \ref{fig:binCNN_loss} and \ref{fig:binCNN_prof}. Model 3 was the most suitable CNN model to classify the binarized fingerprints with an accuracy of $75.94\%$ and an F1-score of $73.99\%$. Interestingly, the comparison between current and previous performance indicates that the approach used to binarize the fingerprints significantly improves the classification process.

\begin{figure}
    \includegraphics[width=\linewidth]{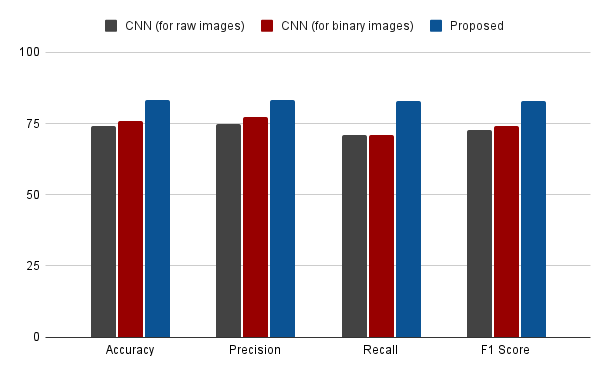}
    \centering
    \captionsetup{justification=centering}
    \caption{Comparing the performance (in \%) of CNN-based deep learning classifiers against the proposed solution.}
    \label{fig:cnnVSproposed}
\end{figure}

At last, we compared the performance of the best 2 CNN-based deep learning models against our proposed fuzzy learning-based solution. Fig. \ref{fig:cnnVSproposed} clearly shows that the proposed solution stands out on all four measures, i.e., Accuracy, Precision, Recall, and F1 Score.

\begin{figure}
    \includegraphics[width=\linewidth]{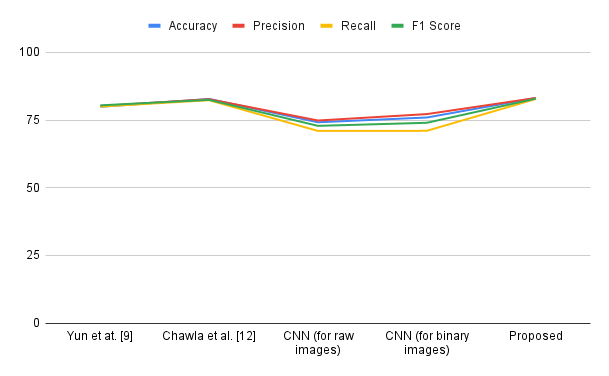}
    \centering
    \captionsetup{justification=centering}
    \caption{Comparing the performance (in \%) of CNN-based deep learning classifiers against different ML-based solutions.}
    \label{fig:dlVSml}
\end{figure}

From Fig. \ref{fig:dlVSml}, another important pattern could be noticed while comparing a few other traditional machine learning solutions against the CNN-based approaches. It shows that the CNN models underperform also when compared with \cite{YUN2006101} and \cite{10.1613}. The unbalanced data input can be a primary reason behind this significant difference in the performance of these models. It also supports that the proposed data balancing method, leveraging KL divergence, plays a crucial role in boosting the data classification quality of the newly introduced fuzzy logic-based classifier. Unfortunately, we could not integrate this data balancing method with CNN as it cannot generate images.

\subsection{Hybrid Fingerprint Orientation Map}
\label{subsub:imageGen}
The proposed hybrid fingerprint orientation map (HFOM) generation method uses a set of $n$ fingerprints detected as $standard$. The challenging part is to choose the most suitable $standard$ fingerprint. We consider $n = 10$ for the experiment purposes. We got the best fingerprint stacks based on the selected algorithm by sequentially sorting classified images.  

\begin{figure}
    \includegraphics[width=7.6cm]{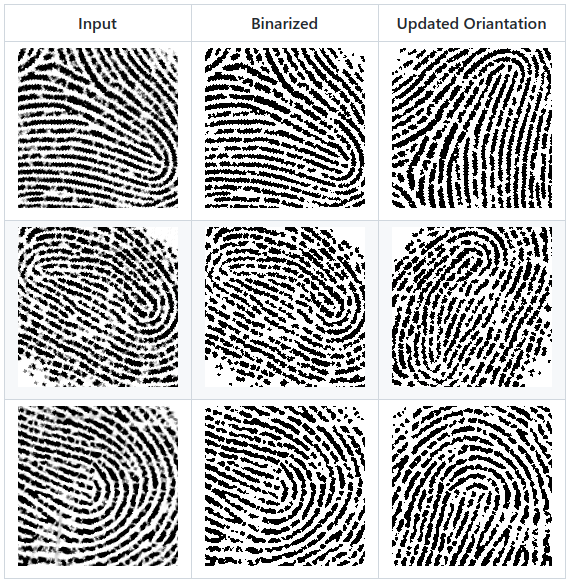}
    \centering
    \captionsetup{justification=centering}
    \caption{Sample fingerprints after binarization and initial orientation change.}
    \label{fig:tillBinPlusOrt}
\end{figure}

The binarization of $n$ best fingerprint images is done after setting the threshold $t = 127$, representing the mid-color intensity between white ($255$) and black ($0$) and generating $F'$. This step is followed by the rotation of $F'_c \in F'$. Combined results are given in Fig. \ref{fig:tillBinPlusOrt}.

\begin{figure}
    \includegraphics[width=7.6cm]{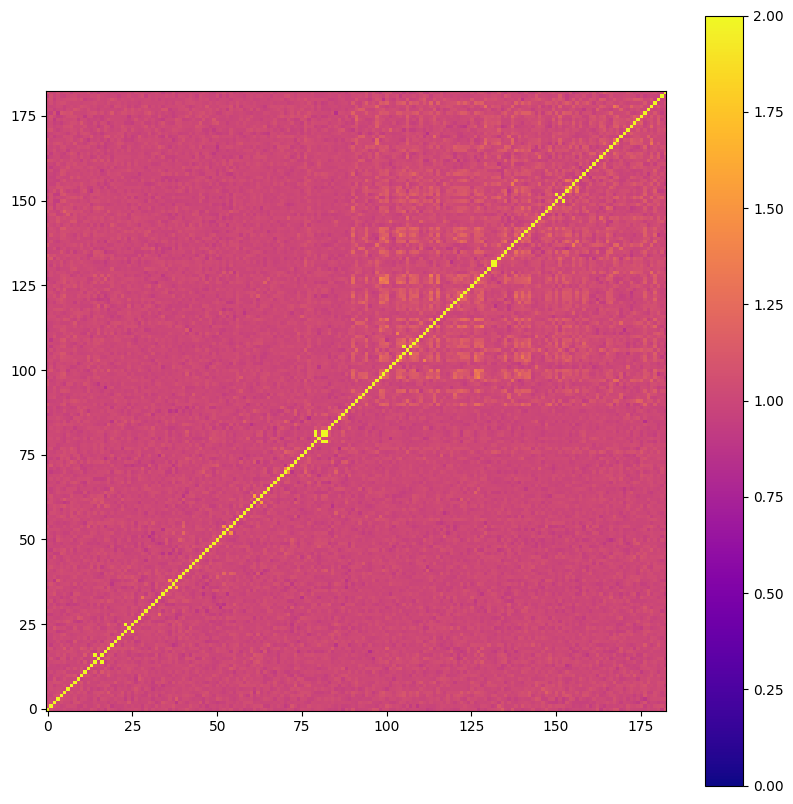}
    \centering
    \captionsetup{justification=centering}
    \caption{SSIM heatmap of \textit{standard fingerprint} image pairs after binarization}
    \label{fig:binNormSSIM}
\end{figure}

\begin{figure}
    \includegraphics[width=7.6cm]{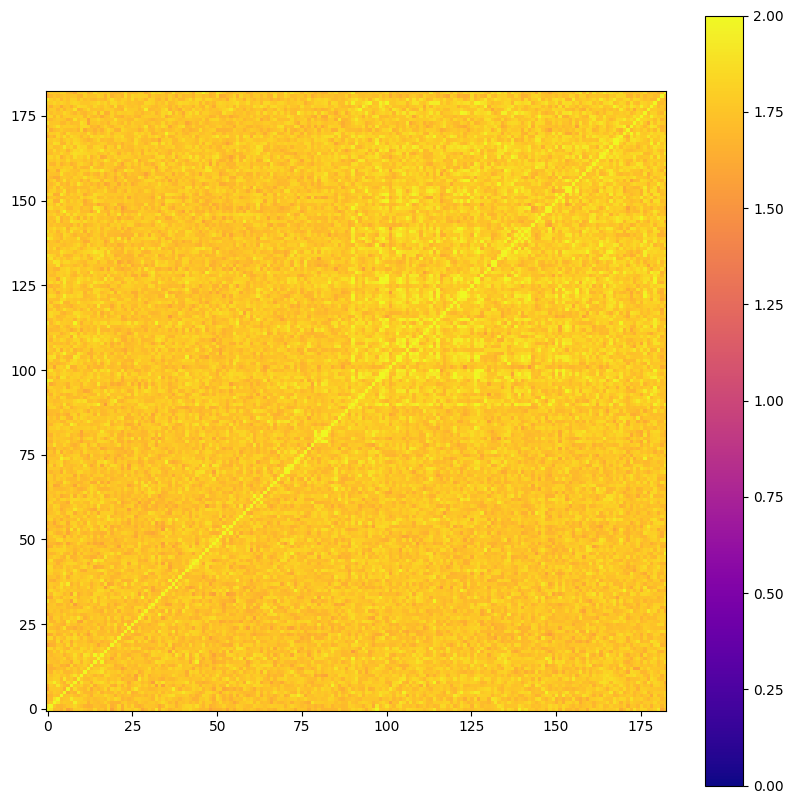}
    \centering
    \captionsetup{justification=centering}
    \caption{SSIM heatmap of $ridge~orientation map$ for \textit{standard fingerprint} image pairs after binarization}
    \label{fig:fomNormSSIM}
\end{figure}

We checked the similarity among the total $560$ images of fingerprints from all three categories from the training dataset. We used the structural similarity (SSIM) for this purpose. The SSIM evaluates the similarities of two images ($I$ and $I'$), represented as $sim(I, I')$ in a range of $[-1, 1]$ that was later converted into a range of $[0,2]$ (Eq. \ref{eq:simNotation}. Here, $-1$ indicates a completely dissimilar pair of, and $1$ signifies a pair of identical images. However, the comparison was not done on the raw images. Instead, we used the corresponding binarized and fingerprint orientation map images. Figure \ref{fig:binNormSSIM} and \ref{fig:fomNormSSIM} show the structural similarities among the set of $183$ images of binarized and proposed orientation maps of fingerprints, respectively.

\begin{equation}
    sim(I,I') = sim(I,I') + 1
    \label{eq:simNotation}
\end{equation}

\begin{figure}
    \includegraphics[width=8cm]{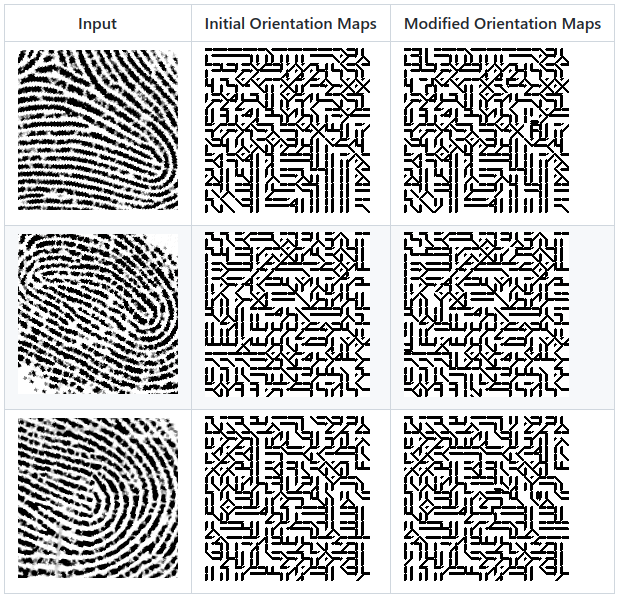}
    \centering
    \captionsetup{justification=centering}
    \caption{Sample fingerprints after initial and modified orientation maps at the block level.}
    \label{fig:tillortMapsFP}
\end{figure}

Before starting with the orientation field generation for the ridge flow in the binarized fingerprints, the images are split into the blocks of an equal square of size $b_{s} \times b_{s}$. We kept $b_{s} = 15$ to have minimal change in the fingerprints with a high increase in the common pixel count. The ridge orientation fields are generated for every $F'_c \in F'$, having the block size $15 \times 15$. In order to maximize the number of common pixels ${P'_{ij}}^c$, the internal blocks are also rotated, and the final modified orientation maps are generated. The results are found in Fig. \ref{fig:tillortMapsFP}.

\begin{figure}
\centering
    \includegraphics[width=4cm]{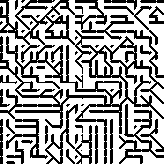}
    \captionsetup{justification=centering}
    \caption{Generated Hybrid Fingerprint Orientation Map (HFOM) $H$.}
    \label{fig:finalHFOM}
\end{figure}

With each of these processes, there has been a subsequent growth notice in standard pixels in Table \ref{tab:commonPxlCount}. The existence of more standard pixels helps generate a more realistic HFOM as it consists of the parts taken from different $\Omega_{c}$. The final HFOM generated is given as Fig. \ref{fig:finalHFOM}.

\begin{table}[h]
\centering
    \begin{tabular}{|l|l|}
    \hline
    \textbf{Step Followed}  &  \textbf{Common Pixels} \\
    \textbf{on $F'_c$}  &  \textbf{on Image Stack}   \\
    \hline \hline
    \multicolumn{1}{|l|}{No Change}  &  76  \\
    \multicolumn{1}{|l|}{Binarization}  &  126  \\
    \multicolumn{1}{|l|}{$F'_c$ Rotation}  &  170  \\
    \multicolumn{1}{|l|}{Ridge Orientation}  & 7736   \\
    \multicolumn{1}{|l|}{Fields Generation}  &    \\
    \multicolumn{1}{|l|}{Orientation Map Modification}  &  8263   \\
    \multicolumn{1}{|l|}{at Block Level}  &     \\
    \hline
    \end{tabular}
    \caption{$P_{count}$ after different steps.}
    \label{tab:commonPxlCount}
\end{table}

\begin{figure}
\centering
    \includegraphics[width=\linewidth]{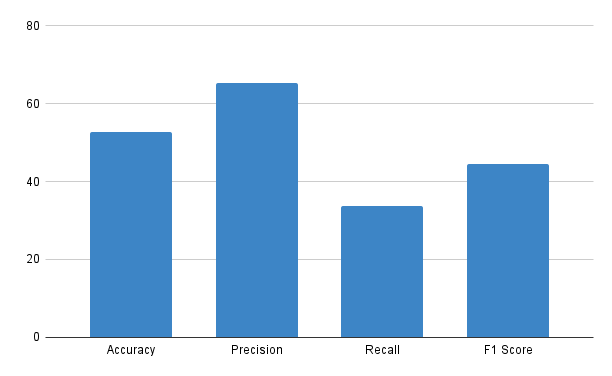}
    \captionsetup{justification=centering}
    \caption{Performance (in \%) of best CNN classifier to detect the FOM image generated from dry, wet, or standard fingerprints.}
    \label{fig:bestHFOM_cnn}
\end{figure}

We tested the efficiency of recognizing the original type of fingerprints among \textit{dry, wet,} and \textit{standard} corresponding fingerprint orientation maps (FOMs) developed using the neural network-based classification models. We again used the architecture mentioned in Section \ref{subsub:CNN} to create six CNN models, which were later trained on several FOMs generated from all the wet, dry, and standard tagged finger impressions, and this time, we found that the model trained with 100 epochs and consisted of 3 subsequent sets of convolution and soft-max layers stood top against the other three. However, as shown in Fig. \ref{fig:bestHFOM_cnn}, the classifier failed to detect the type of origin (dry, wet, standard) of as many as $42.59\%$ of inputted FOM images. It proves the robustness of the proposed approach in processing the fingerprints smartly and highlights the non-tractability of HFOMs for retrieving the original fingerprints. Finally, based on these arguments, we can claim that the proposed HFOM generation process is entirely a one-way method that generates one unique HFOM for the given inputs that could be used as a virtual fingerprint and lays the foundation to design the mechanism for the protection against fingerprint spoofing.


\section{Conclusions}
\label{sec:conclusion}
This paper mainly focuses on developing a hybrid machine learning-based fingerprint classification model with a multi-layered architecture named UC-FLEM capable of working with unbalanced datasets. The proposed model does a multiclass classification with satisfactory results. Following the classification, we introduce a hybrid fingerprint orientation map (HFOM) generation mechanism that uses a set of best-standard fingerprints. 
We have enhanced the binarization technique for fingerprints and found that it plays a good role in the classification while using any classification model, including the neural networks, i.e., CNN. The novel method developed for dataset balancing also significantly improved the classifier's performance. The proposed method uses KL-divergence for the dissimilarities among the datasets. The results have proven that the proposed methods for image binarization, dataset balancing, and classification perform far better than several existing conventional and deep learning-based classification algorithms.

We discussed generating the unique hybrid fingerprint orientation map (HFOM) using the proposed one-way function. To the best of our knowledge, this work is the first of its type where we are generating a kind of visual hash of human fingerprints as HFOM that could be used as a virtual fingerprint. The traditional SSIM method satisfies the claim, which shows that all of the HFOMs have high similarities. Some experiments were conducted to classify HFOMs generated from dry, standard, and wet fingerprints using six distinct Convolutional Neural Network (CNN) classification models varying in architectural configuration. However, these experiments also did not yield any successful results. This provides additional protection to the fingerprint biometrics by providing a proxy for the actual ones and makes it difficult to reconstruct the original fingerprints from the HFOM.

The current threshold does not fit best for binarizing fingerprints with any intensity. Hence, the proposed method has a broad scope of improvement for a more efficient and adaptive binarization technique. Also, neural network-based models can join the orientation fields and generate more realistic HFOM with finer details. 


\bibliographystyle{IEEEtran}
\bibliography{references}

\begin{thebibliography}{10}
\providecommand{\url}[1]{#1}
\csname url@samestyle\endcsname
\providecommand{\newblock}{\relax}
\providecommand{\bibinfo}[2]{#2}
\providecommand{\BIBentrySTDinterwordspacing}{\spaceskip=0pt\relax}
\providecommand{\BIBentryALTinterwordstretchfactor}{4}
\providecommand{\BIBentryALTinterwordspacing}{\spaceskip=\fontdimen2\font plus
\BIBentryALTinterwordstretchfactor\fontdimen3\font minus \fontdimen4\font\relax}
\providecommand{\BIBforeignlanguage}[2]{{%
\expandafter\ifx\csname l@#1\endcsname\relax
\typeout{** WARNING: IEEEtran.bst: No hyphenation pattern has been}%
\typeout{** loaded for the language `#1'. Using the pattern for}%
\typeout{** the default language instead.}%
\else
\language=\csname l@#1\endcsname
\fi
#2}}
\providecommand{\BIBdecl}{\relax}
\BIBdecl

\bibitem{9056799_biometric_app}
H.~Deng, Z.~Qin, Q.~Wu, Z.~Guan, R.~H. Deng, Y.~Wang, and Y.~Zhou, ``Identity-based encryption transformation for flexible sharing of encrypted data in public cloud,'' \emph{IEEE Transactions on Information Forensics and Security}, vol.~15, 2020.

\bibitem{9364994}
Y.~Moolla, A.~De~Kock, G.~Mabuza-Hocquet, C.~S. Ntshangase, N.~Nelufule, and P.~Khanyile, ``Biometric recognition of infants using fingerprint, iris, and ear biometrics,'' \emph{IEEE Access}, vol.~9, 2021.

\bibitem{JAIN201680}
A.~K. Jain, K.~Nandakumar, and A.~Ross, ``50 years of biometric research: Accomplishments, challenges, and opportunities,'' \emph{Pattern Recognition Letters}, vol.~79, 2016.

\bibitem{zimmermann2020password}
V.~Zimmermann and N.~Gerber, ``The password is dead, long live the password--a laboratory study on user perceptions of authentication schemes,'' \emph{International Journal of Human-Computer Studies}, vol. 133, 2020.

\bibitem{manickam2023retraction}
A.~Manickam, E.~Devarasan, G.~Manogaran, M.~K. Priyan, R.~Varatharajan, C.-H. Hsu, and R.~Krishnamoorthi, ``Retraction note: Score level based latent fingerprint enhancement and matching using {SIFT} feature,'' \emph{Multimedia Tools and Applications}, vol.~82, 2023.

\bibitem{YUN2006101}
E.-K. Yun and S.-B. Cho, ``Adaptive fingerprint image enhancement with fingerprint image quality analysis,'' \emph{Image and Vision Computing}, vol.~24, 2006.

\bibitem{manickam2019score}
A.~Manickam, E.~Devarasan, G.~Manogaran, M.~K. Priyan, R.~Varatharajan, C.-H. Hsu, and R.~Krishnamoorthi, ``Score level based latent fingerprint enhancement and matching using {SIFT} feature,'' \emph{Multimedia Tools and Applications}, vol.~78, 2019.

\bibitem{tu2020fingerprint}
Y.~Tu, Z.~Yao, J.~Xu, Y.~Liu, and Z.~Zhang, ``Fingerprint restoration using cubic bezier curve,'' \emph{BMC Bioinformatics}, vol.~21, 2020.

\bibitem{10.1109/ICASSP.2016.7472049}
C.-C. Liao and C.-T. Chiu, ``Fingerprint recognition with ridge features and minutiae on distortion,'' in \emph{ICASSP}, 2016.

\bibitem{7471822}
T.-T. Chu and C.-T. Chiu, ``A cost-effective minutiae disk code for fingerprint recognition and its implementation,'' in \emph{ICASSP}, 2016.

\bibitem{7952518}
H.-R. Su, K.-Y. Chen, W.~J. Wong, and S.-H. Lai, ``A deep learning approach towards pore extraction for high-resolution fingerprint recognition,'' in \emph{ICASSP}, 2017.

\bibitem{shreya2023gan}
S.~Shreya and K.~Chatterjee, ``{GAN}-enable latent fingerprint enhancement model for human identification system,'' \emph{Multimedia Tools and Applications}, 2023.

\bibitem{5456043}
Y.~Jin, W.-S. Soh, and W.-C. Wong, ``Error analysis for fingerprint-based localization,'' \emph{IEEE Communications Letters}, vol.~14, 2010.

\bibitem{9053801}
Y.~Xu, Y.~Wang, J.~Liang, and Y.~Jiang, ``Augmentation data synthesis via {GAN}s: Boosting latent fingerprint reconstruction,'' in \emph{ICASSP}, 2020.

\bibitem{1687-5281}
S.~Bharadwaj, M.~Vatsa, and R.~Singh, ``Biometric quality: a review of fingerprint, iris, and face,'' \emph{EURASIP Journal on Image and Video Processing}, vol.~34, 2014.

\bibitem{10.1613}
N.~Chawla, K.~Bowyer, and W.~Hall, L.O.~Kegelmeyer, ``{SMOTE}: Synthetic minority over-sampling technique,'' \emph{Journal of Artificial Intelligence Research}, vol.~16, 2002.

\bibitem{2018.06.056}
G.~Douzas, F.~Bacao, and F.~Last, ``Improving imbalanced learning through a heuristic oversampling method based on {K}-means and {SMOTE},'' \emph{Information Sciences}, vol. 465, 2018.

\bibitem{4633969}
H.~He, Y.~Bai, E.~A. Garcia, and S.~Li, ``{ADASYN}: Adaptive synthetic sampling approach for imbalanced learning,'' in \emph{IJCNN}, 2008.

\bibitem{10.1111/j.1468-0394.2010.00513.x}
K.~Jackowski and M.~Wozniak, ``Method of classifier selection using the genetic approach,'' \emph{Expert Systems}, vol.~27, 2010.

\bibitem{80959}
W.~Chen, K.~Wong, and J.~Reilly, ``Detection of the number of signals: a predicted eigen-threshold approach,'' \emph{IEEE Transactions on Signal Processing}, vol.~39, 1991.

\bibitem{8489572}
B.~Krawczyk, A.~Cano, and M.~Woźniak, ``Selecting local ensembles for multi-class imbalanced data classification,'' in \emph{IJCNN}, 2018.

\bibitem{momaninovel}
A.~A. Momani and L.~T. K{\'o}czy, ``A novel fingerprint identification fuzzy system using a center-distance weighted local binary pattern,'' in \emph{Computational Intelligence and Mathematics for Tackling Complex Problems 5}, vol. 1127.\hskip 1em plus 0.5em minus 0.4em\relax Springer Nature Switzerland, 2024.

\bibitem{10.1007/978-3-642-23713-3_29}
R.~Arjona, A.~Gersnoviez, and I.~Baturone, ``Fuzzy models for fingerprint description,'' in \emph{Fuzzy Logic and Applications}, vol. 6857.\hskip 1em plus 0.5em minus 0.4em\relax Springer Berlin Heidelberg, 2011.

\bibitem{8461885}
P.-Y. Chou and C.-T. Chiu, ``{PVDC}: A binary descriptor using pore-valley disk code structure for high-resolution partial fingerprint recognition,'' in \emph{ICASSP}, 2018.

\bibitem{10.1007/978-3-319-19324-3_14}
R.~Arjona and I.~Baturone, ``A fingerprint retrieval technique using fuzzy logic-based rules,'' in \emph{Artificial Intelligence and Soft Computing}, vol. 9119.\hskip 1em plus 0.5em minus 0.4em\relax Springer International Publishing, 2015.

\bibitem{10.15837/ijccc.2010.4.2510}
I.~Iancu, N.~Constantinescu, and M.~Colhon, ``Fingerprints identification using a fuzzy logic system,'' \emph{International Journal of Computers}, vol.~5, 2010.

\bibitem{okereafor2020fingerprint}
K.~Okereafor, I.~Ekong, I.~O. Markson, K.~Enwere \emph{et~al.}, ``Fingerprint biometric system hygiene and the risk of {COVID}-19 transmission,'' \emph{JMIR Biomedical Engineering}, vol.~5, 2020.

\bibitem{yang2021biometrics}
W.~Yang, S.~Wang, N.~M. Sahri, N.~M. Karie, M.~Ahmed, and C.~Valli, ``Biometrics for internet-of-things security: {A} review,'' \emph{Sensors}, vol.~21, 2021.

\bibitem{chowdhury2022contactless}
A.~M. Chowdhury and M.~H. Imtiaz, ``Contactless fingerprint recognition using deep learning — {A} systematic review,'' \emph{Journal of Cybersecurity and Privacy}, vol.~2, 2022.

\bibitem{6044711_fp_App}
A.~Kumar and Y.~Zhou, ``Human identification using finger images,'' \emph{IEEE Transactions on Image Processing}, vol.~21, 2012.

\bibitem{castro2023medical}
F.~Castro, D.~Impedovo, and G.~Pirlo, ``A medical image encryption scheme for secure fingerprint-based authenticated transmission,'' \emph{Applied Sciences}, vol.~13, 2023.

\bibitem{9506386_fpGen}
K.~Bahmani, R.~Plesh, P.~Johnson, S.~Schuckers, and T.~Swyka, ``High fidelity fingerprint generation: Quality, uniqueness, and privacy,'' in \emph{ICIP}, 2021.

\bibitem{nilsson2020understanding}
J.~Nilsson and T.~Akenine-M{\"o}ller, ``Understanding {SSIM},'' \emph{CoRR}, vol. abs/2006.13846, 2020.

\bibitem{wang2004image_SSIM_main}
Z.~Wang, A.~C. Bovik, H.~R. Sheikh, and E.~P. Simoncelli, ``Image quality assessment: from error visibility to structural similarity,'' \emph{IEEE Transactions on Image Processing}, vol.~13, 2004.

\bibitem{bakurov2022structural_SSIM_ups}
I.~Bakurov, M.~Buzzelli, R.~Schettini, M.~Castelli, and L.~Vanneschi, ``Structural similarity index ({SSIM}) revisited: {A} data-driven approach,'' \emph{Expert Systems with Applications}, vol. 189, 2022.

\bibitem{5596999_SSIM_FSIM}
A.~Horé and D.~Ziou, ``Image quality metrics: {PSNR} vs. {SSIM},'' in \emph{ICPR}, 2010.

\bibitem{ding2020image_NEW_meth}
K.~Ding, K.~Ma, S.~Wang, and E.~P. Simoncelli, ``Image quality assessment: {U}nifying structure and texture similarity,'' \emph{IEEE Transactions on Pattern Analysis and Machine Intelligence}, vol.~44, 2020.

\bibitem{sampat2009complex_SSIM_etc}
M.~P. Sampat, Z.~Wang, S.~Gupta, A.~C. Bovik, and M.~K. Markey, ``Complex wavelet structural similarity: {A} new image similarity index,'' \emph{IEEE Transactions on Image Processing}, vol.~18, 2009.

\bibitem{fvc2000}
B.~U. of~Bologna, ``{FVC2000} fingerprint verification competition,'' \url{http://bias.csr.unibo.it/fvc2000/db4.asp}, 2000.

\bibitem{10.1023}
L.~Breiman, ``Random forests,'' \emph{Machine Learning}, vol.~45, 2001.

\bibitem{10.1214/aos/1013203451}
J.~H. Friedman, ``Greedy function approximation: A gradient boosting machine.'' \emph{The Annals of Statistics}, vol.~29, 2001.

\bibitem{868688_normCut}
J.~Shi and J.~Malik, ``Normalized cuts and image segmentation,'' \emph{IEEE Transactions on Pattern Analysis and Machine Intelligence}, vol.~22, 2000.

\bibitem{9206836}
A.~G. Medeiros, J.~P.~B. Andrade, P.~B.~S. Serafim, A.~M.~M. Santos, J.~G.~R. Maia, F.~A.~M. Trinta, J.~A.~F. de~Macêdo, P.~P.~R. Filho, and P.~A.~L. Rego, ``A novel approach for automatic enhancement of fingerprint images via deep transfer learning,'' in \emph{IJCNN}, 2020.

\bibitem{6912970_tifs}
M.~Liu, X.~Chen, and X.~Wang, ``Latent fingerprint enhancement via multi-scale patch based sparse representation,'' \emph{IEEE Transactions on Information Forensics and Security}, vol.~10, 2015.

\end{thebibliography}

\end{document}